\documentclass{article}
\usepackage{times}
\usepackage{epsfig}
\usepackage{graphicx}
\usepackage{amsmath}
\usepackage{amssymb}
\usepackage{bbm, dsfont}
\usepackage{color}
\usepackage{cite}
\usepackage[numbers,sort]{natbib}

\usepackage{graphicx}
\usepackage{comment}
\usepackage[margin=1.5in]{geometry}
\usepackage[utf8]{inputenc} % allow utf-8 input
\usepackage{url}            % simple URL typesetting
\usepackage{booktabs}       % professional-quality tables
\usepackage{amsfonts}       % blackboard math symbols
\usepackage{authblk}
 \usepackage{relsize}

\begin{document}
\author{Yiming Xu, Diego Klabjan}

\title{Concept Drift and Covariate Shift Detection Ensemble with Lagged Labels}
\maketitle

\begin{abstract}

In model serving, having one fixed model during the entire often life-long inference process is usually detrimental to model performance, as data distribution evolves over time, resulting in lack of reliability of the model trained on historical data. It is important to detect changes and retrain the model in time. The existing methods generally have three weaknesses: 1) using only classification error rate as signal, 2) assuming ground truth labels are immediately available after features from samples are received and 3) unable to decide what data to use to retrain the model when change occurs. We address the first problem by utilizing six different signals to capture a wide range of characteristics of data, and we address the second problem by allowing \textit{lag of labels}, where labels of corresponding features are received after a lag in time. For the third problem, our proposed method automatically decides what data to use to retrain based on the signals. Extensive experiments on structured and unstructured data for different type of data changes establish that our method consistently outperforms the state-of-the-art methods by a large margin.

\end{abstract}

\section{Introduction}

In most challenging real-world supervised machine learning tasks in model serving such as sentiment analysis of Twitter users and weather forecasting, data evolve and change over time, causing the machine learning models built on historical data to become increasingly unreliable. One reason is usually that the classification or regression models assume stationarity, i.e. the training and test data should be independent and identically distributed \cite{iid1,iid2}. This assumption is often violated, leading to limited generalization ability. The changes of the relationship between features and labels are referred to as a concept drift, while the changes of features only are referred to as covariate shift. When either one occurs, the model performance deteriorates. In order to handle the concept drift or the covariate shift, the easiest way is to retrain the model as soon as a batch of new labeled data is available. This is impractical as 1) continuous retraining is extremely computationally time-consuming and 2) it is a waste of effort when the data distribution does not change. A better strategy is to require the model serving system to continuously diagnose signals such as the classification error and feature reconstruction error, and to automatically adapt to changes in data over time. For example, when the classification error rate increases significantly, the system is supposed to detect the signal and retrain the model.

We consider the model serving process where we continuously receive new batches of data (a batch can as well be a single sample corresponding to streaming) for inference. The data comes without labels which can either arrive immediately after the batch or at any time in the future. Before a new batch arrives, the system needs to make a decision to retrain the model or not and it needs to select a subset of samples to use in retraining if triggered.

There exist several works on drift detection \cite{ddm, eddm, ph, adwin, ewma, cdstudy1} that mainly focus on concept drift with classification performance as the signal for detection. However, this may be problematic. On the one hand, having only one signal may greatly increase the likelihood of false detection or missed detection. On the other hand, drift detection relying on classification performance requires in-time online labeling and is impractical in real-world applications, as in-time online labeling is time-consuming, costly and requires a large amount of human intervention \cite{hi1,hi2, iid2}. 

We address the first problem by including six different signals, capturing different characteristics of data changes such as a lagged classification error rate and model uncertainty. They function as an ensemble and use a tailored majority voting strategy for drift detection, and thus reduce the reliance on one specific signal; the model is also less sensitive to anomaly data. 

Although unavoidable, in order to reduce the reliance on in-time online labeling, we address the second problem by introducing the \textit{lag of labels} setting. In this setting, the system receives the labels of input data after certain time periods to allow time for labeling, instead of receiving labels immediately after a batch as in previous works. In order to detect drift effectively and not waiting until labels are available, our proposed method utilizes a lagged classification error rate for concept drift and other signals for covariate shift, which monitor feature distribution changes as an early indicator of drift. This \textit{lag of labels} scenario is prevalent in real-world applications due to domain expert labeling efforts.

Moreover, most of the existing drift detection algorithms do not have mechanisms to determine what data to use for model retraining when drift is reported, which poses a challenge when applied to real-world applications, because not only do we care about effective and timely drift detection, but the model performance in serving is important as well. Our proposed method automatically determines the data that are used for retraining by collecting samples that are in the warning zone, which can be easily deployed into model serving without human efforts to determine retraining data.

In this work, we propose Concept Drift and Covariate Shift Detection Ensemble (CDCSDE), a drift detection ensemble algorithm in the \textit{lag of labels} setting, where the system receives the labels of input features after a time period due to labeling costs. The ensemble system is composed of six drift detection modules, capturing different characterstics of incoming data such as misclassification rate and feature reconstruction error. The proposed system is also able to decide when to retrain the model and automatically select the data to be used for retraining. We evaluate CDCSDE on both structured and unstructured datasets, and on simulated and real-world datasets. The results show that the proposed method consistently outperforms all the benchmark drift detection models by a large margin.

Our contributions are summarized as follows.
\begin{itemize}
    \item We propose a novel and effective method for drift detection in the \textit{lag of labels} setting.
    \item The proposed method can detect both concept drift and covariate shift; it can determine when to retrain and what data to use to retrain automatically, by utilizing an ensemble of six different drift detectors.
    \item CDCSDE is suitable for both structured and unstructured data.
    \item We conduct extensive experiments on popular drift detection benchmark datasets. The results show the proposed method consistently outperforms all other methods by a large margin.
\end{itemize}

The rest of the paper is organized as follows. In the next section, we provide a review of the relevant literature. In Section 3, we provide the problem description while in Section 4 we discuss the proposed approach in detail. In Section 5, we show the experimental study results. We conclude the paper by reiterating the main contributions in Section 6.

\section{Related Work}

There are several works utilizing statistical control to monitor and detect drift in data streaming. A Cumulative Sum control chart \cite{ph} (CUSUM) is a sequential analysis technique, which is typically used for monitoring change detection. The system reports an alarm as soon as the cumulative sum of incoming data exceeds a user-specified threshold value. The Page Hinkley \cite{ph} (PH) test, a variant of CUSUM, is also a sequential analysis technique often used for change detection in the average of a Gaussian signal. Similar to CUSUM, the PH test alarms a user of a change in the distribution when the test statistic of incoming data is greater than a user-specified threshold. The exponentially weighted moving average \cite{ewma} (EWMA) method monitors the mis-classification rate of a classifier for change detection. It calculates the recent error rate by down-weighting the previous data progressively and reports a drift when the EWMA estimator exceeds an adaptive threshold value. Slightly different from the previous approaches, Adaptive Windowing \cite{adwin} (ADWIN) is an adaptive sliding window algorithm for drift detection, which keeps updated statistics from a window of a variable size. The algorithm takes the binary prediction results for incoming data as input, and decides the size of the window by cutting the statistics window at different points and it analyzes the average of statistics over these sub-windows. Whenever the absolute value of the difference between the two averages from two sub-windows exceeds a threshold, the algorithm concludes that the corresponding expected values are different and reports a drift. The Drift Detection Method \cite{ddm} (DDM) is the most widely used concept drift detection method based on the assumption that the model error rate would decrease as the number of analyzed samples increases, provided that the data distribution is stationary. Same as ADWIN, DDM uses a binomial distribution to describe the model performance. DDM then calculates the sum of the overall classification error and its standard deviation. The most significant difference between DDM and the previous works is that when the sum of the two statistics exceeds a threshold, either drift is detected or the algorithm warns that drift may occur soon. The data that DDM flags as a warning can be potentially used for future retraining.

There are four major differences between our proposed algorithm and the existing works. First, most of the existing works utilize a user-specified threshold for drift detection, and thus the performance of the drift detector largely depends on the choice of the threshold with the model performance being extremely sensitive to it. Our proposed algorithm does not require such threshold. Second, most of the existing works focus on an ideal supervised setting, i.e. they assume the labels are always available as soon as the input features are received, which does not capture many real-world applications due to labeling costs. Our proposed algorithm assumes the \textit{lag of labels} setting, which can report a drift as an early indicator even though the labels are not yet available. Third, these works utilize only a single statistic, e.g. the classification error rate, to monitor changes in data, which suffers from the existence of outliers and several other issues, while our work utilizes an ensemble of six drift detectors that can capture different characteristics of incoming data. Fourth, the previous works only alarm the user about changes in data, omitting deciding the data to be used for retraining. In our work, we progressively select the retraining data based on the calculated warning zone.

\section{Problem Formulation}

A data stream is a data set where observations have time stamps, which induces either a total or a partial order between observations \cite{Webb2018AnalyzingCD}. In our work, we assume classification as the only task, although the proposed method can easily generalize to the unsupervised setting where no labels are available.

Suppose the joint distribution $P(X,Y)$ generates random variables $X$ and $Y$, where $X$ denotes the features for classification and $Y$ denotes the corresponding labels. We further denote $P_t(X,Y)$ as the joint distribution at time $t$. Following the conventional definition, concept drift occurs when

$$P_n(X,Y) \neq P_m(X,Y)$$ for time $n$ and $m$, i.e., the joint distribution changes from time $n$ to time $m$ ($n<m$), which often results in model performance degradation, as the model trained to fit one distribution no longer delivers.

Similarly, covariate shift occurs when

$$P_n(X) \neq P_m(X)$$ for times $n$ and $m$, i.e. the feature distribution has changed. Detection of covariate shift is also of great importance, as on the one hand, it can be used as an early indicator of concept drift, especially when labels are not available or cannot be obtained in-time. On the other hand, most parametric models output probability distributions, which are useful information to learn the model confidence about such predictions. If the model is no longer confident about the predictions, it is desirable to alarm and retrain the model.

In real-world applications, a data stream is usually generated by different joint distributions, as characteristics of incoming data may change over time. If a concept drift or covariate shift occurs, the classification performance is often affected, thus there arises the need of drift detection and subsequent model retraining.

In a data stream in our work, we assume that labels $Y_n$ can only be obtained after a time period $l$, i.e. at time $n$ features $X_n$ are available, while the corresponding labels $Y_n$ are available at time $n+l$. This is a more difficult but practical scenario as opposed to the setting where labels $Y_n$ are available as soon as $X_n$ arrives. In real-world scenarios, the labels may arrive at any time, and thus $l$ may be a random variable following a statistical distribution. 

%In Section 5, we conduct experiments in both cases to validate the proposed method.

%Thus, instead of using a fixed lag of labels in the previous experiments, we further experiment on employing a stochastic lag of labels which follows exponential distribution with scale = 3 and the sampled value is rounded down.

Due to the aforementioned reasons, one cannot have a fixed model during the entire process of model serving, as incoming data distribution might have changed and causes performance degradation. The main tasks are: 1) how to monitor model performance, 2) how to decide if data distribution has changed and 3) what data to use to retrain if change is detected. The ultimate goal is to maximize the progressive accuracy of the provided classifier across all incoming data by addressing the three tasks.

\section{Concept Drift Detection Ensemble and Model Retraining}
In this section, we exhibit our approach to solve the drift detection and model retraining problems.

\begin{figure}[h]
     \includegraphics[width=0.99\textwidth]{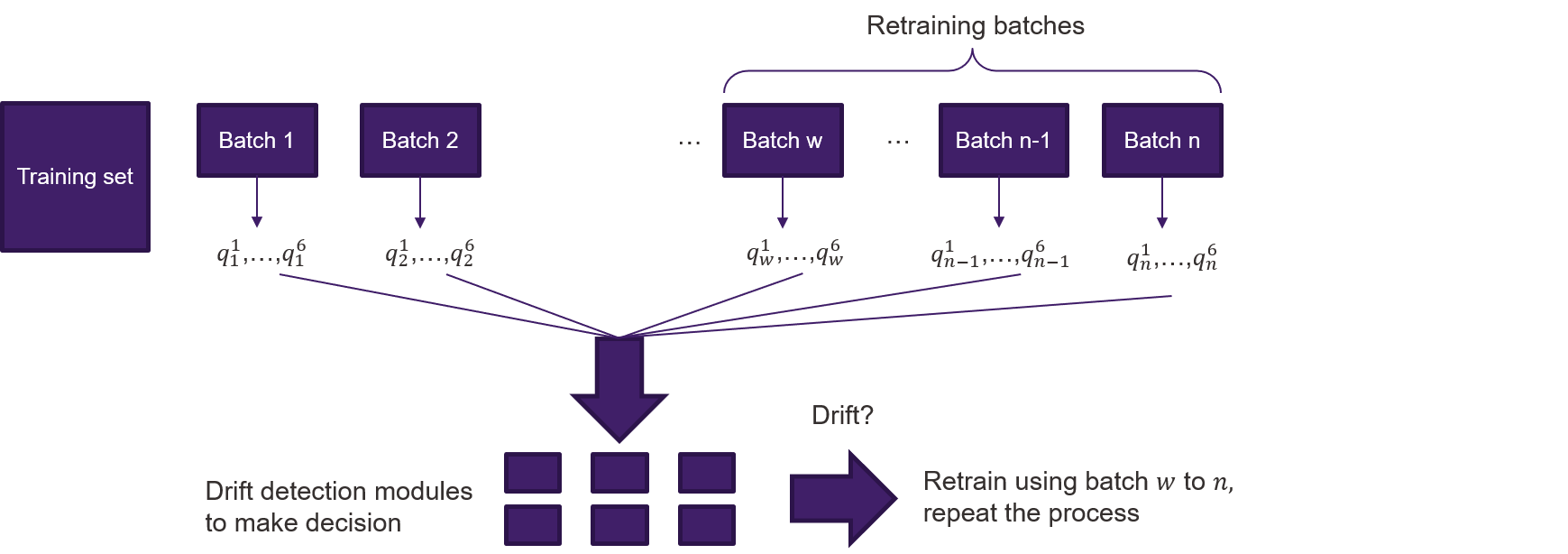}
     \centering
   \caption{Overall approach. For each batch of incoming data, we calculate six descriptive statistics of time series, then utilize drift detection module for each of them to monitor drift and decide what data used to retrain.}
   \centering
   \label{approach_fig}
   \end{figure}

We assume the label lag is $l$. Data $(X_{tr},Y_{tr})$ are used to train the current classification model $clf$. We denote an incoming batch as $B_n=(X_n,Y_n)$, where $Y_n$ are labels for feature set $X_{n-l}$. Sets $Y_1, Y_2, ..., Y_l$ are assumed to be empty.

Figure \ref{approach_fig} shows the general approach of our methodology, which contains six different drift detection modules capturing different characteristics of data: EWMA for delayed KPI, model uncertainty, Hellinger distance, auto-encoder reconstruction error, SPN fitting loss and gradient changes. They are further explained in the following subsections. The combination of the six modules is able to detect both concept drifts and covariate shifts. Based on the tailored majority voting strategy among the six modules, when the system decides to retrain the model, it utilizes proper batches to retrain the model, after which the entire monitoring process is repeated.

\subsection{Descriptive statistics calculation}

We construct six time series which are input to the drift detection modules. At each time step six different scores are calculated, which are either based on only the current batch or based on the current batch as well as the previous batches. The six scores are then added to the six time series for drift detection. We provide details on how we calculate the scores in the following subsections. 

\textbf{EWMA of a delayed classification indicator:}
Let $kpi(Y,X)$ be the most important key performance indicator of $clf$, e.g. error rate or $1-F1$. 
We calculate the exponentially weighted moving average of the delayed KPI as follows. Assume we trace back $k$ batches to calculate the moving average, i.e. at time $n$, we use the KPI of batches $n-k+1$, $n-k+2$,..., $n$, as the labels are delayed by $l$ based on our assumption. For weight decay $w, 0<w<1$, the score is calculated as

\begin{equation}
q^1_n = \mathlarger{\sum}_{i=n-k+1}^{n}kpi(Y_{i-l},clf(X_{i-l})) \cdot w^{n-i+1}.
\end{equation}

\noindent The main difference between (1) and the prior works is that the score defined in (1) avoids using the labels of the current batch due to label delay, and it uses the delayed KPI for drift detection, while the prior works utilizing EWMA assume $l=0$, i.e. they are not designed to handle \textit{lag of labels} and they use a different statistical control method. Despite of this, it generalizes in a straightforward manner to the setting where features and labels arrive simultaneously ($l=0$). Moreover, utilizing EWMA instead of focusing on a single batch is beneficial to the stability of the system, as it is more robust against potential outliers. Intuitively, when EWMA increases significantly, the model suffers from performance degradation and the drift has likely occurred.

\textbf{Model uncertainty:}
It is also helpful to understand how confident the model is regarding the predictions. In the second score, we first predict every sample in the current batch $X_n$ to obtain the predicted probability distributions. For each of these distributions, we construct the histogram of the largest probability and the histogram of the second largest probability, and fit each histogram using a Gaussian distribution to obtain probability density functions $N_1, N_2$. The model uncertainty score is obtained by 

\begin{equation}
q^2_n = \int_{-\infty}^{\infty} min(N_1, N_2).
\end{equation}

\noindent When the overlapping area increases, the mean and variance of the two Gaussians are close to each other, indicating that the largest probability and the second largest probability from predictions are very similar and thus the model is more uncertain regarding the predictions. A significant increase of $q^2$ is a signal for drift, since when the model predictions are uncertain, the likelihood of a change in distribution is high. On the other hand, when the model is uncertain regarding a distribution, even if the predictions are correct (i.e. $q^1$ does not vary much), it is still detrimental to the overall process of model serving, as a small fluctuation of the distribution may move the model decision boundary and makes the model vulnerable to future outliers.

\textbf{Hellinger distance:}
In statistics and measure theory, the Hellinger distance is often used to quantify similarity between two distributions, and is also widely used in tasks such as anomaly detection and classification. We utilize the Hellinger distance to construct the distance between two datasets. Having discretized the features, we define the Hellinger distance between the training set and the current batch as

\begin{equation}
q_n^3=\frac{1}{|F|}\mathlarger{\sum}_{f\in F} \sqrt{\mathlarger{\sum}_{z \in f}( \sqrt{\frac{|X_{tr,f=z}|}{|X_{tr}|}} - \sqrt{\frac{|X_{n,f=z}|}{|X_{n}|}})^2},
\end{equation}

\noindent where $F$ denotes the set of all features. By averaging over all features, we calculate the distance between two datasets or batches. If the distance between the training dataset (i.e. the dataset that the model is fitted on) and the current batch is large, the existing model no longer fits the current distribution and needs to be retrained. Note that if the input data is unstructured such as image data, we calculate the Hellinger distance based on encoded feature vectors from an auto-encoder, instead of the raw features themselves.

\textbf{Auto-encoder reconstruction error:}
Auto-encoders are widely used in tasks such as dimension reduction and embedding learning. In our work, we employ an auto-encoder to measure how different the two datasets or batches are. We first train an auto-encoder using training features $X_{tr}$ and obtain the training reconstruction MSE loss $L_{tr}=MSE(X_{tr},AE(X_{tr}))$. For the current batch $X_n$, we calculate the test reconstruction loss $L_n=MSE(X_n,AE(X_n))$. The auto-encoder reconstruction score is defined as

\begin{equation}
q_n^4=tanh(\frac{L_{te}}{L_{tr}}).
\end{equation}

\noindent This score allows us to measure divergence by how large the test reconstruction error is compared to the training error. The increase of $q^4$ indicates that the auto-encoder is unable to fully reconstruct the incoming data and thus the covariate shift has occurred.

\textbf{Sum-Product Networks: }
The Sum-Product Network (SPN) is a deep probabilistic model widely used as a black box density estimator by comparing the likelihoods on tasks such as image completion and image classification \cite{Vergari2018VisualizingAU}. Similarly as for $q^4$, we use an SPN to monitor if the incoming distribution has changed or not, compared to the training data. After training an SPN using training data $X_{tr}$, for batch $X_n$, we obtain the log of  negative log-likelihood $log(-ll_n)$. We consider the log since log likelihood from SPN is very small. The SPN module score is defined as

\begin{equation}
q_n^5=log(-ll_n).
\end{equation}

\noindent As vanilla SPNs do not take unstructured data as input, in such cases, we feed the embedded features from the encoder of the auto-encoder to SPN. The higher the score is, the less likely $X_n$ is generated by the same distribution as $X_{tr}$. 

\textbf{Gradient changes:}
Proposed in \cite{kungangwork, kungangworkarxiv}, the changes of gradients can also be used to tackle the drift detection task, i.e. the larger the gradient changes, the more likely the data has changed. Instead of using the conventional gradients, we utilize natural gradients instead, which show promising performance in areas such as robotics and control. As natural gradients are rescaled by the Fisher information matrix, they are more stable and often used to solve issues such as catastrophic forgetting. 

Let the optimal parameters from the training data $(X_{tr},Y_{tr})$ be $\theta ^*$. We evaluate the natural gradients at time $n$ using batch $(X_{n-l},Y_{n})$  as $\nabla _N Loss = \nabla Loss(X_{n-l},Y_{n},\theta^*) \cdot F^{-1}$, where $Loss$ is the conventional loss function (cross-entropy in classification) and $F$ is the Fisher information matrix, approximated by using Kronecker factorization \cite{kfac}. The gradients change score is obtained by

\begin{equation}
q_n^6=\frac{(\nabla_N Loss)^T (\nabla_N Loss)}{u}
\end{equation}
where $u$ denotes the dimension of $\theta^*$.
\noindent As gradient changes, the score should increase which is a signal for potential drifts. 

It is worth pointing out that \cite{kungangwork, kungangworkarxiv} also transformed the gradient by the inverse Fisher information matrix in a manner that is quite similar, but not identical, to our method. Namely, \cite{kungangwork, kungangworkarxiv} monitored the components of the `decoupled' score vector (i.e. the gradient transformed by the inverse Fisher information matrix). Their statistic also implicitly transformed the score vector by the inverse of the covariance matrix, which is closely related to the Fisher information matrix.

\subsection{Drift detection}

Each descriptive statistic defined in the previous subsection generates a time series over time. The remaining problem is how to detect drift based on these time series. We describe next for the current batch $X_n$, the resulting algorithm to report drift, warning or safe signals, which utilizes an ensemble of six independent drift detection modules to monitor drift. 

Similarly to \cite{ddm}, to detect drift of a time series $\{q_n\}_{n=1,2,...}$, we denote the progressive average by $p_i=\mathlarger{\sum}^i_{n=1} q_n/i$ and standard deviation by $s_i=std(p_1,p_2,...,p_i)$ . We also denote the minimum values at time $i$ by $p_{min}^i$ and $s_{min}^i$ such that $p_{min}^i + s_{min}^i = \displaystyle\min_{1\leq n\leq i}(p_n+s_n)$. Note that different from \cite{ddm}, at each time step we receive a batch instead of a single observation, thus our time series is not generated by Bernoulli distributions. To monitor drift, we apply the following rules:

If $p_i+s_i>p^i_{min}+2\cdot s^i_{min}$, the system is in the warning zone;

If $p_i+s_i>p^i_{min}+3\cdot s^i_{min}$, we report drift and retrain the model using batches in the warning zone;

If $p_i+s_i<p^i_{min}+2\cdot s^i_{min}$, the system exits the warning zone and is safe from drift.

There are two assumptions for this statistical control module. First, the values in the time series decrease when the data distribution is stationary, and thus the system should either exit the warning zone or stay in the safe zone. Second, a significant increase in the time series indicates the existence of a drift.

With a simple drift detection module, we would be able to detect drift and decide what data to use to retrain based on a single time series. Having six different time series, our method CDCSDE works as an ensemble. To detect a drift event, we employ a tailored majority voting rule as follows: if the drift detection module on $q^1$ reports drift, then the system reports drift and retrains all models (classifier, auto-encoder and SPN), as the KPI is the ultimate goal; otherwise, if most of the remaining modules (i.e. not less than 3 modules) report a drift, CDCSDE reports drift and retrains all models. 

If drift is reported, the data used to retrain is determined by CDCSDE as well. Suppose the latest warning zone (if a module exits the warning zone, then the previous warning zone is not included in future retraining) for each module is $W_i,i=1,2,...,6$. The union of these warning batches $\bigcup\limits_{i=1}^6 W_i$ is used to retrain all models.

\section{Experimental Results}
In this section, we report experiments on both simulated and real-world datasets to evaluate the proposed method with the KPI being accuracy. The simulated datasets have ground truth about drift. For simulated datasets, 5 different metrics are used: the mean accuracy value over all batches (MA), mean time between false alarms (MTFA), mean time to detection (MTD), missed detection rate (MDR), total number of drifts detected (TD). A good drift detection method should have high MA and MTFA (or no MTFA due to no false alarms) as well as low MTD and MDR. Low MTD is desired because a good algorithm should be able to detect drift sooner instead of waiting for a long time to react (i.e. smaller time to detect drift). 

Low TD is also beneficial since it reduces retraining time and computational costs. Among these metrics, MA is the most important as in real-world model serving the overall model performance is what we care most about. In real-world datasets only MA and TD are available as we do not have the knowledge when drift `indeed' occurs. We conduct experiments on both structured and unstructured data to validate the effectiveness of CDCSDE.

We compare the proposed method with four so-far-best-performing benchmarks: PH, ADWIN, EWMA and DDM, which are discussed in Section 2. For fair comparison, the benchmarks use the delayed KPI as signal in the \textit{lag of labels} setting and the same training data (i.e. the first 50 batches) are used across all methods. We experimented with other values of the number of initial batches to consider for training to find out that the performance is insensitive to this number. The number of batches 50 leads to 3,200 samples which is an appropriate number for both structured and unstructured data sets under consideration. In our experiments we also observe that with our retraining strategy, most of the time the number of retraining batches are less than 50. Moreover, in our observation, the number of retraining batches does not show an increasing or decreasing trend when additional batches arrive, since our algorithm selects only the useful batches to retrain instead of including as many batches as possible.

In the following experiments, the \textit{lag of labels} is assumed to follow the exponential distribution and thus the time between events is a Poisson process, i.e., a process where events occur continuously and independently at a constant rate. We employ scale = 4 and a sampled value is rounded down to an integer. We also study the sensitivity with respect to the \textit{lag of labels} and we construct an ablation study.

\subsection{Structured data stream}

We first perform experiments on structured data which consist of the following widely used datasets.

\textbf{1. Sea} \cite{sea}: simulated data with abrupt drift, 50,000 samples, 3 features and 2 classes. There are 3 drifts in total, each occuring after 12,500 samples. \textbf{2. Sine2} \cite{ddm}: simulated data with abrupt drift, 100,000 samples, 2 features and 2 classes. There are 9 drifts in total, each occuring after 10,000 samples.  \textbf{3. Elec} \cite{elec}: real-world data, 45,312 samples, 8 features and 2 classes, which are recorded every half an hour for two years from the Australian New South Wales Electricity Market. \textbf{4. Weather} \cite{weather}: real-world data, 18,159 samples, 8 features and 2 classes, which record weather measurements from over 7,000 weather stations worldwide to provide a wide scope of weather trends. \textbf{5. Temp} \cite{temp}: real-world data, 4,137 samples, 24 features and 2 classes, which are collected from a monitor system mounted in houses.

Across all datasets, we utilize a 2-layer auto-encoder with shape `Input-6 neurons-Reconstructed Input' and a one-layer feed-forward network with shape `6-Output' as the classifier which takes the embedded features from the auto-encoder as input. Also taking the embedded features as input, the SPN consists of 4 layers, i.e. normal, product, sum, product. We optimize the models with the Adam optimizer with the 0.001 initial learning rate. The batch size is set as 64.

\begin{table}[h!]
\centering
%\footnotesize
\caption{Results on simulated structured datasets. } 
  \begin{tabular}{l| c | c| c | c | c | c| c | c| c|c |c}
  \hline
  
  & \multicolumn{5}{c|}{\textbf{Sea}} & \multicolumn{5}{c|}{\textbf{Sine2}} \\ \hline
      \textbf{Method}&\textbf{MA}&\textbf{MTFA}&\textbf{MTD}&\textbf{MDR}&\textbf{TD}&\textbf{MA}&\textbf{MTFA}&\textbf{MTD}&\textbf{MDR}&\textbf{TD}&\textbf{AVG}\\ \hline
PH  &69.50& \textbf{-}& 43.5&33.3 & 3 & 69.78& 38.0 & 22.3 & 33.3 & 8&69.64 \\
ADWIN & 77.59& 9.5 &18.0 & 0 &14 & 78.39& 15.9 & 40.0 & 33.3 &16& 77.99 \\ 
EWMA & 75.73 & 15.5 & 22.0 & 0& 7&47.92& 31.5 & 25.0 & 44.4&11&61.83 \\
DDM & 85.32 & \textbf{-}& 42.5 & 33.3 & 2&77.38& \textbf{-} & \textbf{9.3} & 55.6& 5&81.35 \\ \hline
CDCSDE & \textbf{88.08} &\textbf{-} &\textbf{17.7} &\textbf{0} &3 & \textbf{83.69} & \textbf{-}& 10.4 &\textbf{22.2} & 8&\textbf{85.89} \\ \hline
  \end{tabular}
  \label{tab:simulated_structured}
\end{table}

\begin{table}[h!]
\centering
\caption{Results on real-world structured datasets.} 
  \begin{tabular}{l| c | c| c|c | c| c|c  }
  \hline
  
  & \multicolumn{2}{c|}{\textbf{Elec}} & \multicolumn{2}{c|}{\textbf{Weather}} & \multicolumn{2}{c|}{\textbf{Temp}} \\ \hline
 \textbf{Method}&\textbf{MA}&\textbf{TD}&\textbf{MA}&\textbf{TD}&\textbf{MA}&\textbf{TD}&\textbf{AVG}\\ \hline
PH  & 57.51& 6 & 56.18& 2& 67.00&6 &60.23\\
ADWIN & 58.38 &2 & 67.70& 1&78.45 &2&68.18 \\ 
EWMA & 59.61 &8 &32.29& 1&74.39 &3&55.46\\
DDM & 58.63 &1 & 67.70& 1& 66.25&1&64.19\\ \hline
CDCSDE & \textbf{67.71} &5 &\textbf{74.97} &2 &\textbf{82.80} &2&\textbf{75.16}\\ \hline
  \end{tabular}
  \label{tab:real_structured}
\end{table}

From Table \ref{tab:simulated_structured} where `-' denotes not available, i.e. no false alarms or only one false alarm, `AVG' denotes the average MA across all datasets, and the numbers in bold denote the best across all methods if applicable, we observe that among all evaluation metrics, CDCSDE consistently outperforms the benchmarks, especially in the most important metric MA, as it effectively detects drift and decides what data to use to retrain. Our relative improvements on average accuracy are 23.33\%, 10.13\%, 38.91\%, 5.58\% for PH, ADWIN, EWMA and DDM, respectively. Regarding MTFA, CDCSDE predicts no false detection in Sea and only one false detection in Sine2, while other benchmarks do not exhibit a stable performance, resulting in more false alarms and high MTFA. Our method also achieves the best performance in MTD due to the timely alarm of drift, with 17.7 on Sea and 10.4 on Sine2, while other benchmarks generally being less sensitive to the drifts. Lastly, among all methods, CDCSDE achieves the smallest MDR and a reasonable TD. 

We also conduct experiments on Elec, Weather and Temp datasets to provide better insights on how our method performs on real datasets. The results are provided in Table \ref{tab:real_structured} where `-' denotes not available, i.e. no false alarms or only one false alarm, `AVG' denotes the average MA across all datasets, and the numbers in bold denote the best across all methods if applicable. Note that we do not have measures such as MTFA, MTD, etc. as they are real-world datasets without the ground truth information regarding drifts. From Table \ref{tab:real_structured}, CDCSDE consistently achieves the best performance in MA for all benchmarks, which again validates the effectiveness of CDCSDE. Our relative improvements on average accuracy are 24.79\%, 10.24\%, 35.52\%, 17.09\% for PH, ADWIN, EWMA and DDM, respectively. The TD of our method is also reasonable, while some other benchmarks either report too many drifts or do not report a drift at all.

\subsection{Unstructured data stream}

In order to evaluate our method on unstructured data which is more ubiquitous in real-world applications, we conduct experiments utilizing conventional image classification datasets MNIST and USPS \cite{USPS}. They contain the same 10 digits as labels (i.e. 0-9), but their distributions differ. The two datasets are widely used in domain adaptation tasks due to a moderate domain gap. In our work, we employ them to create 5 different data streams to validate CDCSDE. 

\begin{figure}[h]
     \includegraphics[width=1.1\textwidth]{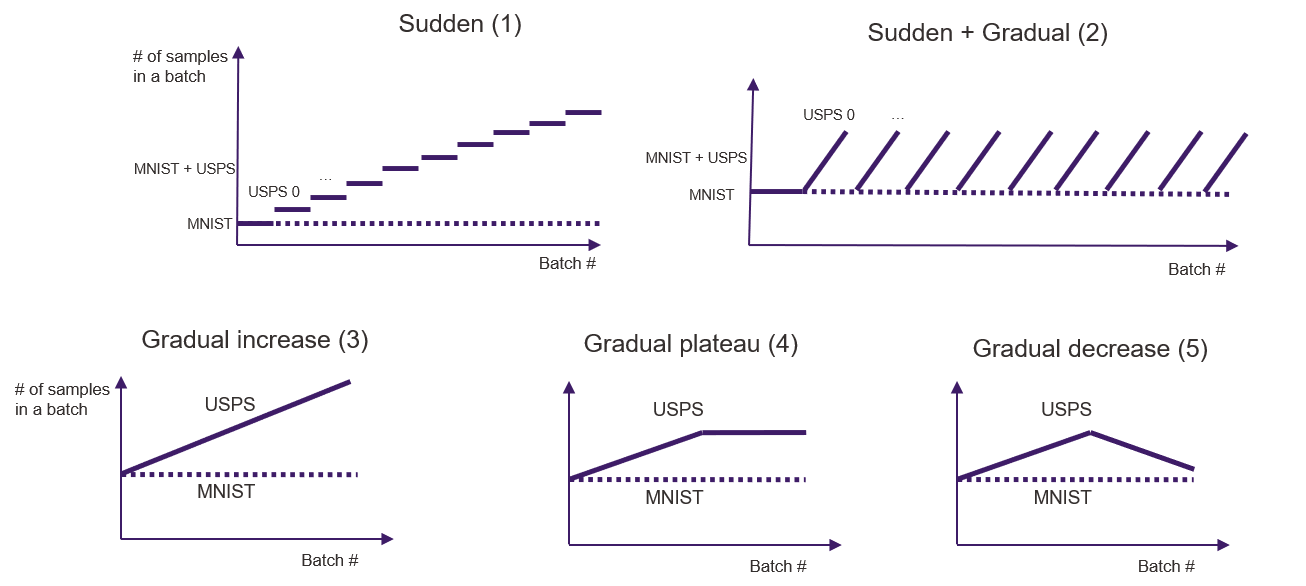}
     \centering
   \caption{Unstructured datasets with various types of drifts.}
   \centering
   \label{uns_case}
   \end{figure}
   
Across all scenarios, we use MNIST as the `base' dataset, i.e. all incoming batches are supposed to contain MNIST 0-9. USPS is used as the `drift' dataset, i.e. samples from USPS are added to the data stream in a specific way to introduce drifts. The batch size is set as 64 and the total number of samples is 60,000 (the size of the MNIST training set), and thus the number of batches is 60,000/64 = 938. The details of each dataset are provided as follows, see also Figure \ref{uns_case}. There are 9,298 USPS samples in total. 

\textbf{1. Sudden drift:} 
Samples of one digit from USPS are introduced after each 100 batches. Therefore, there are  9 sudden drifts in total. \textbf{2. Sudden and gradual drift:}
Samples of one digit from USPS are gradually introduced after each 100 batches. There are  9 sudden drifts in total. \textbf{3. Gradual drift - increase:}
USPS 0-9 digits are gradually and increasingly introduced. \textbf{4. Gradual drift - plateau:}
USPS 0-9 digits are gradually introduced. After the $500^{th}$ batch , the rate of adding USPS 0-9 digits is kept unchanged. \textbf{5. Gradual drift - decrease:}
USPS 0-9 digits are gradually introduced. After the $500^{th}$ batch , USPS 0-9 digits are gradually removed at the same rate of introducing.

\begin{table}[h!]
\centering
%\footnotesize
\caption{Results on unstructured datasets with sudden drifts.} 
  \begin{tabular}{l| c | c| c | c | c | c| c | c| c|c |c}
  \hline
  
  & \multicolumn{5}{c|}{\textbf{Sudden}} & \multicolumn{5}{c|}{\textbf{Sudden and gradual}} \\ \hline
      \textbf{Method}&\textbf{MA}&\textbf{MTFA}&\textbf{MTD}&\textbf{MDR}&\textbf{TD}&\textbf{MA}&\textbf{MTFA}&\textbf{MTD}&\textbf{MDR}&\textbf{TD}&\textbf{AVG}\\ \hline
PH  &71.59& \textbf{-}& 52.0&77.8 & 2 & 73.48& \textbf{-} & 43.0 & 88.9 & 2 &72.53\\
ADWIN & 85.49& 29.3 &\textbf{23.4} & 22.2 &19 & 84.17&29.4 &31.9 & 44.4 &24&84.83 \\ 
EWMA & 90.92 & 53.9 & 29.4 & 33.3& 11 & 92.01& 49.2 & 26.1 & 33.3&19&90.47 \\
DDM & 87.00 & \textbf{-}& 25.5 & 77.8 & 2&85.51& \textbf{-} & 35.5 & 77.8& 2 &86.26\\ \hline
CDCSDE & \textbf{97.41} &252.5 &26.3 &\textbf{11.1} &10 & \textbf{97.89} & 178.3&\textbf{18.3} &\textbf{11.1} & 11 &\textbf{97.65}\\ \hline
  \end{tabular}
  \label{tab:sudden_unstructured}
\end{table}

Across all datasets, we utilize a 2-layer CNN with output shape 500 as the encoder and a one-layer feed-forward network with shape `500-10' as classifier which takes the embedded features from the encoder as input. Also taking the embedded features as input, the SPN consists of 4 layers, i.e. normal, product, sum, product. We optimize our models by using the Adam optimizer with 0.001 initial learning rate. The batch size is 64.

We first experiment on the sudden drift, and sudden and gradual drift scenarios. From Table \ref{tab:sudden_unstructured}, we observe that when sudden drift occurs, the conventional benchmarks generally suffer from high MDR (MDR of PH and DDM are as high as 88.9\% and 77.8\%) due to the lack of ability for detecting changes in feature distributions, as they only take the classification error rate as input. CDCSDE achieves 11.1\% MDR, i.e. only 1 drift missed, which shows the ability of the proposed method for detecting both covariate shifts and concept drifts. CDCSDE's TD are 10 and 11 for the two scenarios which is reasonable for a total of 9 drifts, while other benchmarks either report too many drifts (24 for ADWIN) or too few drifts (2 for PH and DDM). Similarly, the MTFA metric of CDCSDE is much higher than the benchmarks when their TD is reasonable (as otherwise there would be no false alarms at all), indicating that the proposed method is much less likely to make false alarms. As a comparison, ADWIN reports a false alarm every 29.3 time steps on average. The MTD metric of the proposed method is 26.3 for the sudden dataset and 18.3 for the sudden and gradual dataset which is the lowest among all benchmarks, which again shows the ability of detecting drifts as low MTD indicates the method can respond to the changes in data distributions in time. Most importantly, the MA of CDCSDE consistently outperforms other benchmarks by a large margin due to the accurate and timely detection of drifts. Our relative improvements on average accuracy are 34.63\%, 15.11\%, 7.94\%, 13.20\% with respect to PH, ADWIN, EWMA and DDM, respectively.

\begin{table}[h!]
\centering
\caption{Results on unstructured datasets with gradual drifts.} 
  \begin{tabular}{l| c | c| c|c | c| c|c  }
  \hline
  
  & \multicolumn{2}{c|}{\textbf{Gradual increase}} & \multicolumn{2}{c|}{\textbf{Gradual plateau}} & \multicolumn{2}{c|}{\textbf{Gradual decrease}}\\ \hline
 \textbf{Method}&\textbf{MA}&\textbf{TD}&\textbf{MA}&\textbf{TD}&\textbf{MA}&\textbf{TD} & \textbf{AVG}\\ \hline
PH  & 85.63& 3 & 81.28& 2& 81.12&4&82.68 \\
ADWIN & 89.45 &13 & 91.65& 9&87.17 &7 &89.42\\ 
EWMA & 91.11 &11 &88.32& 9&92.68 &10&90.70\\
DDM & 92.31 &2 & 94.15& 3& 93.68&3&93.38\\ \hline
CDCSDE & \textbf{97.99} &3 &\textbf{97.76} &4 &\textbf{97.50} &3&\textbf{97.75}\\ \hline
  \end{tabular}
  \label{tab:gradual_unstructured}
\end{table}

We then conduct experiments only with gradual drifts to observe model performance if there are no sudden drifts in the data stream, the results shown in Table \ref{tab:gradual_unstructured}. From Table \ref{tab:gradual_unstructured} we observe a relatively low TD for CDCSDE and the MA greatly exceeds the benchmarks across all three datasets, which shows the robustness of the proposed method for gradual drift as well. Our relative improvements on average accuracy are 18.23\%,9.31\%, 7.72\%, 4.68\% with respect to PH, ADWIN, EWMA and DDM, respectively.

\textbf{Ablation study}
In order to establish how important each component of CDCSDE is, we conduct an ablation study on three unstructured datasets: sudden, sudden and gradual, gradual decrease. The mean accuracies are shown in Table \ref{tab:ablations_add} and Table \ref{tab:ablations_wo}.

\begin{table}[h!]
\centering
%\footnotesize
\caption{Ablation study: increasingly adding components. For example, + Hellinger denotes the model with components EWMA error rate, uncertainty and Hellinger. }
  \begin{tabular}{r | r  | r | r |r}\hline
       & \textbf{Sudden} & \textbf{Sudden and gradual} & \textbf{Gradual decrease}&\textbf{AVG}\\ \hline
       EWMA error rate& 94.38 &95.68 & 92.18&94.08 \\\hline
+ uncertainty & 95.19 & 95.88 &93.91& 94.79\\\hline
+ Hellinger & 95.59 &95.70& 94.24& 95.18\\\hline
+ AE error& 96.10 & 96.30 &95.31& 95.90\\\hline
+ SPN likelihood& 97.25& 97.10 &96.24& 96.86\\\hline
+ gradient norm& 97.41 & 97.89 &97.50& 97.60\\\hline
  \end{tabular}
  \label{tab:ablations_add}
\end{table}

We first add each component cumulatively at one time to observe performance changes reported in Table \ref{tab:ablations_add}. The largest gains are from the SPN and gradient norm modules, which empirically shows that monitoring changes in feature distribution is beneficial to model performance.

\begin{table}[h!]
\centering
%\footnotesize
\caption{Ablation study: MA decrease without each component. For example, w/o uncertainty denotes the entire model without the uncertainty component.}
  \begin{tabular}{r | r  | r | r |r}\hline
       & \textbf{Sudden} & \textbf{Sudden and gradual} & \textbf{Gradual decrease}&\textbf{AVG}\\ \hline
w/o EWMA error rate& \textbf{2.61} &\textbf{3.59} & \textbf{2.33}&\textbf{2.84} \\\hline
w/o uncertainty & 0.11 & 0.37 &0.30& 0.26\\\hline
w/o Hellinger & 0.21 &0.28& 0.17& 0.22\\\hline
w/o AE error& 0.60 & 0.88 &0.91& 0.80\\\hline
w/o SPN likelihood& 1.38& 1.60 &1.10& 1.36\\\hline
w/o gradient norm& 1.01 & 1.25 &1.07& 1.11\\\hline
  \end{tabular}
  \label{tab:ablations_wo}
\end{table}

We then experiment removing each component to observe the accuracy decrease in Table \ref{tab:ablations_wo}, which shows similar patterns as Table \ref{tab:ablations_add}. The model without SPN suffers from the second largest performance drop, which again validates that modeling feature distribution as an early indicator of drift is important. It is not surprising that CDCSDE without EWMA's error rate suffers from the largest performance degradation, as it utilizes both labels and features to calculate accuracy of predictions and a good accuracy in model serving is what we are ultimately interested in.

\textbf{Varying the level of \textit{lag of labels}}
We also experiment how the model performs on different levels of \textit{lag of labels}. In addition to $l\sim exp(scale=4)$ as in previous experiments, we conduct experiments on small lag $l\sim exp(2)$ and large lag $l\sim exp(10)$. The results are shown in Table \ref{tab:differentlag}.

\begin{table}[h!]

\centering
%\footnotesize
\caption{Results on unstructured datasets with different levels of \textit{lag of labels}.} 
  \begin{tabular}{l| c | c| c | c | c | c| c | c| c|c |c}
  \hline
  
  & \multicolumn{5}{c|}{\textbf{Sudden}} & \multicolumn{5}{c|}{\textbf{Sudden and gradual}} \\ \hline
      \textbf{Method}&\textbf{MA}&\textbf{MTFA}&\textbf{MTD}&\textbf{MDR}&\textbf{TD}&\textbf{MA}&\textbf{MTFA}&\textbf{MTD}&\textbf{MDR}&\textbf{TD}&\textbf{AVG}\\ \hline \hline

 & \multicolumn{10}{c}{$l\sim exp(2)$ }\\ \hline
 
PH  &73.94& \textbf{-}& 45.5&77.8 & 3 & 73.01& \textbf{-} & 39.0 & 77.8 & 3 &73.48\\
ADWIN & 87.74& 25.5 &22.0 & 33.3 &17 & 90.01& 29.2 &\textbf{21.2}& 33.3 &16&88.38 \\ 
EWMA & 92.01 & 59.0 & \textbf{19.1} & 33.3& 12 & 90.19& 29.9 & 26.8 & 44.4&14&90.60 \\
DDM & 87.64 & 39.0& 32.0 & 88.9 & 3&87.10& \textbf{-} & 30.0 & 77.8& 2 &86.26\\ \hline
CDCSDE & \textbf{97.57} &287.0 &24.0 &\textbf{11.1} &10 & \textbf{97.71} &151.0&21.5 &\textbf{11.1} & 10 &\textbf{97.65}\\ \hline\hline

 & \multicolumn{10}{c}{$l\sim exp(4)$ } \\ \hline
PH  &71.59& \textbf{-}& 52.0&77.8 & 2 & 73.48& \textbf{-} & 43.0 & 88.9 & 2 &72.53\\
ADWIN & 88.10& 31.6 &\textbf{25.1} & 33.3 &19 & 84.41& 28.4 &29.5 & 55.6 &14&86.26 \\ 
EWMA & 93.11 & 65.5 & 28.5 & 33.3& 8 & 90.82& 45.0 & 20.2 & 33.3&11&91.97 \\
DDM & 87.00 & \textbf{-}& 25.5 & 77.8 & 2&85.51& \textbf{-} & 35.0 & 77.8& 2 &86.26\\ \hline
CDCSDE & \textbf{97.21} &252.5 &26.3 &\textbf{11.1} &10 & \textbf{97.49} & 178.3&\textbf{18.3} &\textbf{11.1} & 11 &\textbf{97.35}\\ \hline \hline

 & \multicolumn{10}{c}{$l\sim exp(10)$ } \\ \hline
PH  &69.70& 34.0& 41.0&88.9 & 3 & 71.40& 56.0 & 51.0 & 77.8 & 4 &70.55\\
ADWIN &87.45& 29.8 &32.0 & 33.3 &22 & 86.30& 27.0 &33.5 & 44.4 &17&86.88 \\ 
EWMA & 86.98 & 55.3 & 39.5 & 44.4& 10 & 89.33& 58.2 & 28.3 & 44.4&13&88.16 \\
DDM & 84.57 & 33.5& 45.5 & 77.8 & 4&82.05& 40.0 & 36.0 & 77.8& 4 &83.31\\ \hline
CDCSDE & \textbf{96.01} &\textbf{191.0} &\textbf{28.6} &\textbf{22.2} &11 & \textbf{95.41} &\textbf{117.0}&\textbf{24.9} &\textbf{11.1} & 11 &\textbf{95.71}\\ \hline
  \end{tabular}
  \label{tab:differentlag}

\end{table}

From Table \ref{tab:differentlag}, we observe that in general models perform better for smaller \textit{lag of labels}, which is as expected because when the labels arrive sooner, such up-to-date information can be immediately taken into account. When the labeling costs are high, i.e. it takes more time for labels to arrive, the system can only use delayed labels to detect drifts. Comparing among all methods, CDCSDE still shows stable performance across all metrics, consistently outperforming benchmarks by a considerable margin for all levels of \textit{lag of labels}.

\textbf{Fixed \textit{lag of labels}}
Instead of using a stochastic \textit{lag of labels} as in the previous experiments, we further experiment on employing a fixed \textit{lag of labels} of $l=2,4$ and $10$. The results are shown in Table \ref{tab:randomlag}.

\begin{table}[h!]

\centering
%\footnotesize
\caption{Results on unstructured datasets with fixed \textit{lag of labels}. } 
  \begin{tabular}{l| c | c| c | c | c | c| c | c| c|c |c}
  \hline
  
  & \multicolumn{5}{c|}{\textbf{Sudden}} & \multicolumn{5}{c|}{\textbf{Sudden and gradual}} \\ \hline
      \textbf{Method}&\textbf{MA}&\textbf{MTFA}&\textbf{MTD}&\textbf{MDR}&\textbf{TD}&\textbf{MA}&\textbf{MTFA}&\textbf{MTD}&\textbf{MDR}&\textbf{TD}&\textbf{AVG}\\ \hline \hline

 & \multicolumn{10}{c}{$l=2$ }\\ \hline
        
PH  &72.40& 18.5 & 41.5 &77.8 & 4 & 75.41& \textbf{-} & 31.0 & 77.8 & 3 &73.91\\
ADWIN & 91.21& 53.5 &\textbf{23.5} & 33.3 &17 & 88.10& 28.9 &30.1 & 33.3 &14&89.66 \\ 
EWMA & 90.01 & 41.0 & 32.3 &44.4& 12 & 90.83 & 35.2 & 23.6 & 44.4&10&90.42 \\
DDM & 89.02 & 54.5& 28.5 & 77.8 & 4&87.45&38.5& 29.0 & 88.9& 3 &88.24\\ \hline
CDCSDE & \textbf{97.73} &\textbf{-} &25.1 &\textbf{22.2} &8 & \textbf{97.49} &\textbf{-}&\textbf{20.9} &\textbf{11.1} & 9 &\textbf{97.61}\\ \hline\hline

 & \multicolumn{10}{c}{$l=4$ } \\ \hline
PH  &71.04& \textbf{-}& 55.5&77.8 & 3 & 73.01& \textbf{-} & 50.0 & 77.8 & 3 &72.03\\
ADWIN & 87.39& 36.0 &33.8 & 33.3 &14 & 88.50& 31.5 &44.6 & 44.4 &13&87.95 \\ 
EWMA & 89.59& 35.5 & 39.3 & 44.4& 12 & 88.19& 40.0 & 27.5 & 44.4&13&88.89 \\
DDM & 86.24 & 39.0& 32.0 & 88.9 & 3&84.20& \textbf{-} & 30.0 & 77.8& 2 &86.26\\ \hline
CDCSDE & \textbf{97.27} &225.0 &\textbf{30.1} &\textbf{11.1} &10 & \textbf{97.51} &151.0&\textbf{22.5} &\textbf{11.1} & 10 &\textbf{97.39}\\ \hline\hline

 & \multicolumn{10}{c}{$l=10$ } \\ \hline
PH  &70.01& 21.5& 51.0&77.8 & 4 & 71.21& 41.0 & 71.5 & 77.8 & 5 &70.61\\
ADWIN & 84.10& 25.8 &\textbf{33.0} & 55.6 &18& 83.51& 35.8 &55.5 & 44.4 &19&83.81 \\ 
EWMA & 85.19 & 49.0 & 41.0 & 55.6& 8 & 83.91& 48.8 & \textbf{28.3} & 44.4&23&84.55 \\
DDM & 83.87 & 41.0& 50.3 & 66.7 & 5&81.29& 51.3 & 36.5 & 77.8& 5 &82.58\\ \hline
CDCSDE & \textbf{96.51} &\textbf{170.0} &34.1 &\textbf{22.2} &9 & \textbf{95.80} &\textbf{108.5}&29.7 &\textbf{33.3} & 8 &\textbf{96.16}\\ \hline
  \end{tabular}
  \label{tab:randomlag}
\end{table}

Compared with Table \ref{tab:sudden_unstructured}, we observe similar patterns in Table \ref{tab:randomlag}. The benchmarks generally are either too sensitive to drifts or unable to detect drifts. CDCSDE still shows robust performance across all metrics, consistently outperforming benchmarks by a considerable margin even in the fixed \textit{lag of labels} scenario. As expected, Table \ref{tab:randomlag} reflects the intuition that constant lag yields higher accuracy. The differences are not substantial, but they are consistent. Observations are similar regarding other metrics: for example, TD in the fixed \textit{lag of labels} scenario is smaller than the stochastic scenario with reasonable MDR.

\section{Conclusions}

In this paper, we present a novel and effective drift detection method in the practical \textit{lag of labels} setting, which is able to detect both the concept drift and covariate shift and automatically decide what data to use to retrain, with the help of the ensemble of different drift detectors. Extensive experiments on structured and unstructured data for different type of drifts have shown that our method consistently outperforms the state-of-the-art drift detection methods by a large margin.

\bibliography{ref}
\bibliographystyle{splncs04}

\end{document}